\documentclass[10pt,twocolumn,letterpaper]{article}

\usepackage{cvpr}
\usepackage{times}
\usepackage{epsfig}
\usepackage{graphicx}
\usepackage{amsmath}
\usepackage{amssymb}
\usepackage{multirow}
\usepackage{threeparttable}
\usepackage{verbatim}

\usepackage[breaklinks=true,bookmarks=false]{hyperref}

\cvprfinalcopy 

\def\cvprPaperID{****} 

\begin{document}

\title{Data-Free Point Cloud Network for 3D Face Recognition}

\author{Ziyu Zhang, Feipeng Da\thanks{Corresponding Author}, Yi Yu\\
School of Automation, Southeast University\\
Key Laboratory of Measurement and Control of Complex Systems of Engineering\\
{\tt\small \{zzy1993,dafp,yuyi\}@seu.edu.cn}
}

\maketitle

\begin{abstract}
   Point clouds-based Networks have achieved great attention in 3D object classification, segmentation and indoor scene semantic parsing.
   In terms of face recognition, 3D face recognition method which directly consume point clouds as input is still under study. Two main factors account for this: One is how to get discriminative face representations from 3D point clouds using deep network; the other is the lack of large 3D training dataset.
   To address these problems, a data-free 3D face recognition method is proposed only using synthesized unreal data from statistical 3D Morphable Model to train a deep point cloud network. To ease the inconsistent distribution between model data and real faces, different point sampling methods are used in train and test phase.
   In this paper, we propose a curvature-aware point sampling(CPS) strategy replacing the original furthest point sampling(FPS) to hierarchically down-sample feature-sensitive points which are crucial to pass and aggregate features deeply.
   A PointNet++ like Network is used to extract face features directly from point clouds.
   The experimental results show that the network trained on generated data generalizes well for real 3D faces.
   Fine tuning on a small part of FRGCv2.0 and Bosphorus, which include real faces in different poses and expressions, further improves recognition accuracy.
   
\end{abstract}

\section{Introduction}
Recently Deep learning approach achieved promising results in face recognition and related applications.
Face recognition task has transformed from simple constrained frontal face to more challenge tasks, such as pose-invariant, expression-invariant, age-invariant face recognition. However, CNNs-based 3D face recognition has just raised. Most 3D face recognition systems\cite{Huibin2014Expression,Lei2016A,Li2015Towards,mian2007efficient,Spreeuwers2011Fast} focus on traditional methods, which extract hand-crafted features for verification or identification task. But this methods extremely rely on face alignment algorithm and feature descriptors, which limit the scalability as compared to the end-to-end learning approaches.

Two challenges for 3D face recognition are how to design a face point clouds-aware deep network and the lack of large 3D face database to train.
The successful face recognition networks are mostly operated on images\cite{parkhi2015VGGface,Schroff_2015facenet,Zulqarnain_Gilani_2018}. Convolution filters are applied to extract deep and discriminative features for identification and verification. These networks nearly get perfect results in some dataset\cite{Learned2016LFW,cao2018vggface2,yi2014CASIA}.
However, the most common data format for 3D face is mesh or point clouds whose the non-grid structure results in common 2D networks cannot work directly.
One alternative method is to project point cloud data into range map\cite{Soltani_2017_CVPR,kim2017deep}, multi-view image\cite{Jian2016Research,Feng_2018_CVPR} or voxelize to volumetric representation\cite{jackson2017large,Qi_2016_CVPR} so as to adapt for 2D networks.
However, these data reduction methods will lead to information loss or data structure misaligned\cite{wei20193d}. For example, range maps only include the depth in $z$ coordinate neglects the $xy$ information. Volumetric representation introduces quantization errors during voxelization of the point cloud. Besides, volumetric is memory inefficient.
We propose to directly consume raw point clouds to network which helps to dig 3D representations without loss information. PointNet\cite{charles2017pointnet:} structure was the first one to use a symmetric function to aggregate point-wise and order-invariant features on raw point clouds. However, PointNet only aggregated global feature and neglected the local characters which is very importance for 3D shape representation. PointNet++\cite{qi2017pointnet++:} applies PointNet hierarchically to capture both local and global features for object recognition and segmentation on point clouds. We focus more on face features and a curvature-aware point sampling method is proposed to filter facial points and can be easily integrated into PointNet++. A modified PointNet++ is used to extract face features directly from raw face point clouds.

Another challenge is the lack of large 3D face database. As the deep learning is data greedy, it needs adequate data to show its advantage.
In 3D object classification task, ModelNet40 dataset\cite{wu2015modelnet} is most used to train and evaluate. It contains 12311 CAD scans from 40 classes, average 300 scans each class.
In 2D face recognition, CASIA Webface dataset\cite{yi2014CASIA} including 0.5 million images for 10k identities is usually used for tiny study.
However, in 3D face, the largest public dataset now is the LS3DFace\cite{Zulqarnain_Gilani_2018}, which is composed of 1853 identities totally 31860 face scans. The available datasets are still far from training the 3D face network.
So, we need to generate more 3D data for training. Gilani and Mian\cite{Zulqarnain_Gilani_2018} proposed to synthesize new identities from models or by interpolating. But these methods need computational preprocessing such as dense correspondence or registration. And the generate face looks same in expression.
We propose a data-free method for 3D face recognition that we only use generated data from Gaussian Process Morphable Models(GPMM)\cite{luthi2017GPMM} in training and real faces in testing.
The advantage of GPMM is that we can generate unlimited number of faces by shape and expression coefficients. But we must take care of the distribution gap between generated data and real 3D faces.
We constrain the face area to mimic the train data distribution in the test phase.

\begin{figure*}[]
	\begin{center}
		\includegraphics[width=1\linewidth]{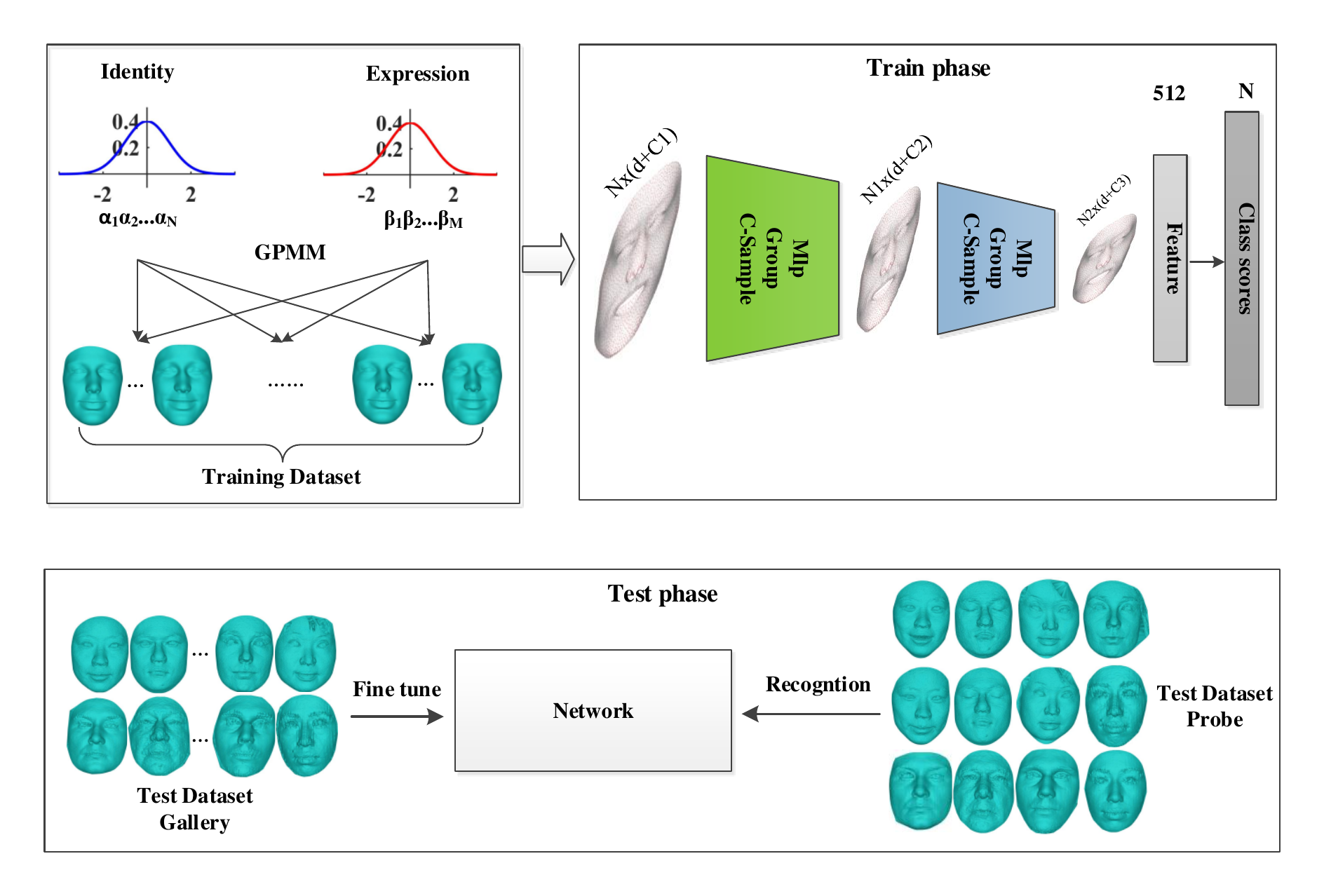}

	\caption{An overview of proposed 3D face point clouds recognition framework.
		Training dataset consists of different 3D scans with expressions generated by GPMM. In train phase, the generated faces pass through two feature abstract layers including proposed curvature-aware point sampler, neighbors grouper and multi-layer perceptrons(MLP) to extract feature embedding for classification task. In the test phase, the feature embedding directly extracted from real faces can be used for face verification or identification. Alternatively, fine tuning on the gallery set can further improve the feature discriminative.
	}
	\label{fig:overview}
	\end{center}
\end{figure*}
An overview of our proposed 3D face recognition method is illustrated in Fig.~\ref{fig:overview}.
Main modules for face recognition includes training dataset generating, Network training and testing. In data generating, GPMM is used to synthesize large number 3D faces. During train phase, the network is treated as a classifier with softmax. While in testing phase, we first fine tune on a few real faces and then achieve face verification and identification using the 512-dims features.
We evaluate our proposed method on FRGCv2.0\cite{phillips2005FRGCv2} and Bosphorus\cite{savran2008bosphorus} datasets with pose and expression disturbance.

The main contributions of this work are as follows:
\begin{itemize}
\item We propose to use GPMM to generate large training data and compensate the distribution difference between the generated data and real faces by constrain the face sampling area.
\item We introduce the point clouds for 3D face recognition and a face curvature-aware sampling strategy is integrated into PointNet++ to localize key face points and enhance the feature more discriminative \footnote{The code is available in github.com/alfredtorres/3DFacePointCloudNet}.
\item Comprehensive experiments are conducted on 3D face datasets with pose and expressions to demonstrate that our point cloud face recognition framework is competitive and has potential to solve high-level face recognition problems in 3D domain.
\end{itemize}

\section{Related work}
\label{Sec:Releatedwork}
\subsection{3D face recognition} 
Most previous methods\cite{berretti2013matching,drira20133d,gupta2010anthropometric,Yuyi2019Sparse} works directly on the 3D space of facial scans, since full information of the raw data is kept on this original domain. An overview of 3D face recognition method is presented in\cite{abate20072d,bowyer2006a,Patil20153}. Existing 3D methods can be largely classified into two categories: local or global feature descriptor methods. local descriptor based methods match local 3D point signatures derived from the curvatures, shape index or normals. For instance, Gupta\cite{gupta2010anthropometric} matched the 3D Euclidean and geodesic distances between pairs of 25 anthropometric fiducial landmarks to perform 3D face recognition. Berretti \etal~\cite{berretti2013matching} represented a 3D face with multiple mesh-DOG keypoints and local geometric histogram descriptors while Drira \etal~\cite{drira20133d} represented the facial surface by radial curves emanating from the nosetip.

As alternatives, some other methods\cite{kim2017deep,Zulqarnain_Gilani_2018} find reasonable mappings from the 3D space to canonical 2D domains. Kim \etal~\cite{kim2017deep} proposed to frontalize a 3D scan, generate 2.5D depth map and extract the depth map's features by VGG16 network to represent the 3D face. Gilani and Mian\cite{Zulqarnain_Gilani_2018} proposed a method that projects face point cloud into depth, azimuth and elevation maps to generate a three channel image. These methods all reduce the face representation dimension from 3D point cloud to 2D images, so as to use CNNs to extract deep features. Gilani and Mian\cite{Zulqarnain_Gilani_2018} projected two more maps than Kim\cite{kim2017deep} so as to get more representative feature. So if we project more channels to form a 4-channels, 5-channels, ...... multi-channels map, we can loss less 3D information and obtain more discriminative feature.
\subsection{Facial data generation}
One of the main factors for the nearly perfect performance in 2D face recognition is the ability of CNNs to learn from massive training data. For instance, FaceNet\cite{Schroff_2015facenet} was trained on a private database containing 200M labeled images of 8M identities while VGG-Face\cite{parkhi2015VGGface} used 2.6M faces of 2622 people. The public used 2D test database LFW\cite{Learned2016LFW} contains more than 13,000 images of faces collected from the web. And the CAISIA webface\cite{yi2014CASIA} including 0.5 million of 10k identities and the VGGFace2\cite{cao2018vggface2} including 3.3 million of 9.1k persons are commonly used for training. However, 3D face scans need special sensors and technique to obtain which limit the 3D face recognition network. For example, the FRGCv2\cite{phillips2005FRGCv2} database contains only 4,007 3D scans of 466 persons while the Bosphorus\cite{savran2008bosphorus} contains 4,666 scans of 105 persons. The largest publicly available 3D face dataset, ND-2006\cite{faltemier2007ND-2006} (a superset of FRGCv2) has only 13,540 scans of 888 unique identities and took over two years to collect.(see in Tab~\ref{tab:Database})
\begin{table}[h]
\caption{Comparison of 2D and 3D face databases.} 
	\label{tab:Database}
	\begin{center}
		\small
		\begin{tabular}{c|c|c|c}
			\hline
			\multicolumn{1}{l|}{Modality} & Name  & Identity & Image/Scan \\ \hline
			\multirow{4}{*}{2D}          & LFW\cite{Learned2016LFW}       & 5749       & 13233      \\
			& CASIA\cite{yi2014CASIA}     & 10k        & 0.5M       \\
			& VGGFacev2\cite{cao2018vggface2} & 9.1k       & 3.3M       \\
			& MS1Mv2    & 85k        & 3.9M       \\ \hline
			\multirow{6}{*}{3D}          & FRGCv2\cite{phillips2005FRGCv2}    & 466        & 4007       \\
			& BU3DFE\cite{yin2006BU3DFE}    & 100        & 2500       \\
			& Bosphorus\cite{savran2008bosphorus} & 105        & 4666       \\
			& CASIA-3D\cite{xu2006casia-3d}  & 123        & 4674       \\
			& ND-2006\cite{faltemier2007ND-2006}   & 422        & 9443       \\
			& LS3DFace\cite{Zulqarnain_Gilani_2018}  & 1853       & 31860      \\ \hline
		\end{tabular}
	\end{center}
\end{table}

Some works propose to reconstruct faces from 2D face images\cite{Yi_2019_MMFace,Gecer_2019_GANFIT} or create synthetic faces from existing face models\cite{Zulqarnain_Gilani_2018}. Blanz and Vetter\cite{blanz19993DMM} proposed a 3D face morphable model (3DMM) using a multiple PCA-based linear subspace method to represent the facial identity and expression. Dou \etal~\cite{Dou_2017_CVPR} reconstruct 3DMM parameters with deep neural network. By means of a mutil-task and a fusion convolutional network, the reconstructed face has more impressive expressions. Yi \etal~\cite{Yi_2019_MMFace} propose to address the face reconstruction in the wild by using a multi-metric regression network to align a 3DMM to an input image. Gecer \etal~\cite{Gecer_2019_GANFIT}takes a radically different approach and harness the power of GANs and DCNNs in order to reconstruct the facial texture and shape from single images. GANs are utilized to train a very powerful generator of facial texture in UV space and then optimize the parameters with the supervision of pretrained deep identity features through an end-to-end differentiable framework.
Gilani and Mian \cite{Zulqarnain_Gilani_2018} proposed a technique to generate a training dataset of 3.1M scans of 100K identities by simultaneously interpolating between the facial identity and facial expression spaces.
\subsection{Deep learning on point clouds} 
Recently we see a surge of interest in designing deep learning architectures suited for point clouds\cite{qi2017pointnet++:,charles2017pointnet:,li2018pointcnn,wang2018SpecGNN,wang2019dgcnn,xu2018spidercnn,wu2019pointconv}, which demonstrated remarkable performance in 3D object classification and segmentation. The difficulty to transform the deep networks from 2D image to 3D point cloud is to design a well sufficient operation or convolution to the unordered data which should be invariant to the orders. 
PointNet\cite{charles2017pointnet:} started the use of deep learning networks for 3D point cloud processing. PointNet used a symmetric function to aggregate point-wise feature which is invariant to the permutations of input points. 
PointNet++\cite{qi2017pointnet++:} hierarchical applied PointNet recursively to obtain both global and local features on point cloud. Besides,  PointCNN\cite{li2018pointcnn} proposed to learn $X$-transform to allow a convolution operator to work directly on point cloud data. 
SpiderCNN\cite{xu2018spidercnn} comprised of SpiderConv units extends convolutional operations from regular grids to irregular point sets that can be embedded in $\mathbb{R}^n$, by parametrizing a family of convolutional filters. 
DGCNN\cite{wang2019dgcnn} and SpecGCN\cite{wang2018SpecGNN} introduces Graph CNN into the point cloud processing. 
In DGCNN, EdgeConv operator is just a convolution-like operations to capture local geometric structure by construct a local neighborhood graph. In addition, the graph is updated by each layer using the nearest neighbors in the feature space, making the respect field larger.
In SpecGCN, graph convolution is carried out on a nearest neighbor graph constructed from a point’s neighborhood to jointly learn features. And a novel pooling strategy to aggregate information from within clusters of nodes that are close to one another in their spectral coordinates is proposed, leading to richer overall feature descriptors.
However, these methods have so far only been used for object classification and segmentation. To the best of our knowledge, no 3D point cloud based face recognition technique that leverages deep learning-based embeddings to match two 3D face has been developed yet.

\section{Face augmentation and point cloud recognition network}
\label{Sec:Main}
Given a pair of face point clouds, the goal of our method is to verify whether these two scans belong to one identity or not. The face point cloud is represented as a set of seven-dimensional points $ F_i = [x_p, y_p, z_p, n_{xp}, n_{yp}, n_{zp}, c_p] $, where $i = 1,...,N, p = 1,..,P$, $N$ is the number of faces. $P$ is the number of points each face and $P$ varies in real faces which adds difficult to face correspondence. We consider the $xyz$ coordinate as the face shape and use normal vectors extract deep features. $c$ is curvature obtained according to .
\subsection{3D face augmentation}
We propose to synthesize face images by sampling from a statistical 3D Morphable Model of face shape and expression. The generator can synthesize an arbitrary amount of facial identities with different expressions. We assume that the facial identity is fully determined by the 3D face shape. We use the GPMM \cite{luthi2017GPMM}, for which the shape distribution is estimated from 200 neutral high-resolution 3D face scans. The parameters follow a Gaussian distribution.
The identity face model consist of shape model and expression model
\begin{equation}
M_s=(\mu_ s,\sigma _s, U_s) , M_e=(\mu_ e,\sigma _e, U_e)
\end{equation}
where $\mu _{\{s,t\}}\in \mathbb{R}^{3m}$ are the mean, $\sigma _{\{s,t\}}\in \mathbb{R}^{n-1}$ the standard deviations and $U _{\{s,t\}}=[u_1, ...u_n]\in \mathbb{R}^{3m\times n-1}$ are an orthogonal basis of principle components of shape and expression.
New faces are generated from the model as linear combinations of the principal components
\begin{equation}
\left\{\begin{matrix}
s(\alpha ) = \mu _s + U_sdiag(\sigma  _s)\alpha\\ 
e(\beta ) = \mu _e + U_ediag(\sigma  _e)\beta
\end{matrix}\right.
\end{equation}
The coefficients $\alpha,\beta$ are independent and normally distributed with unit variance under the assumption of normally distributed training examples and a correct mean estimation.

By drawing random samples from this distribution we generate random 3D face meshes with unique shape $\alpha_i$ and expression $\beta_j$.
A new face instance $F$ is represented as
\begin{equation}F = s(\alpha_i) + e(\beta_j)\end{equation}

\begin{figure}[t]
	\begin{center}
		\includegraphics[width=1\linewidth]{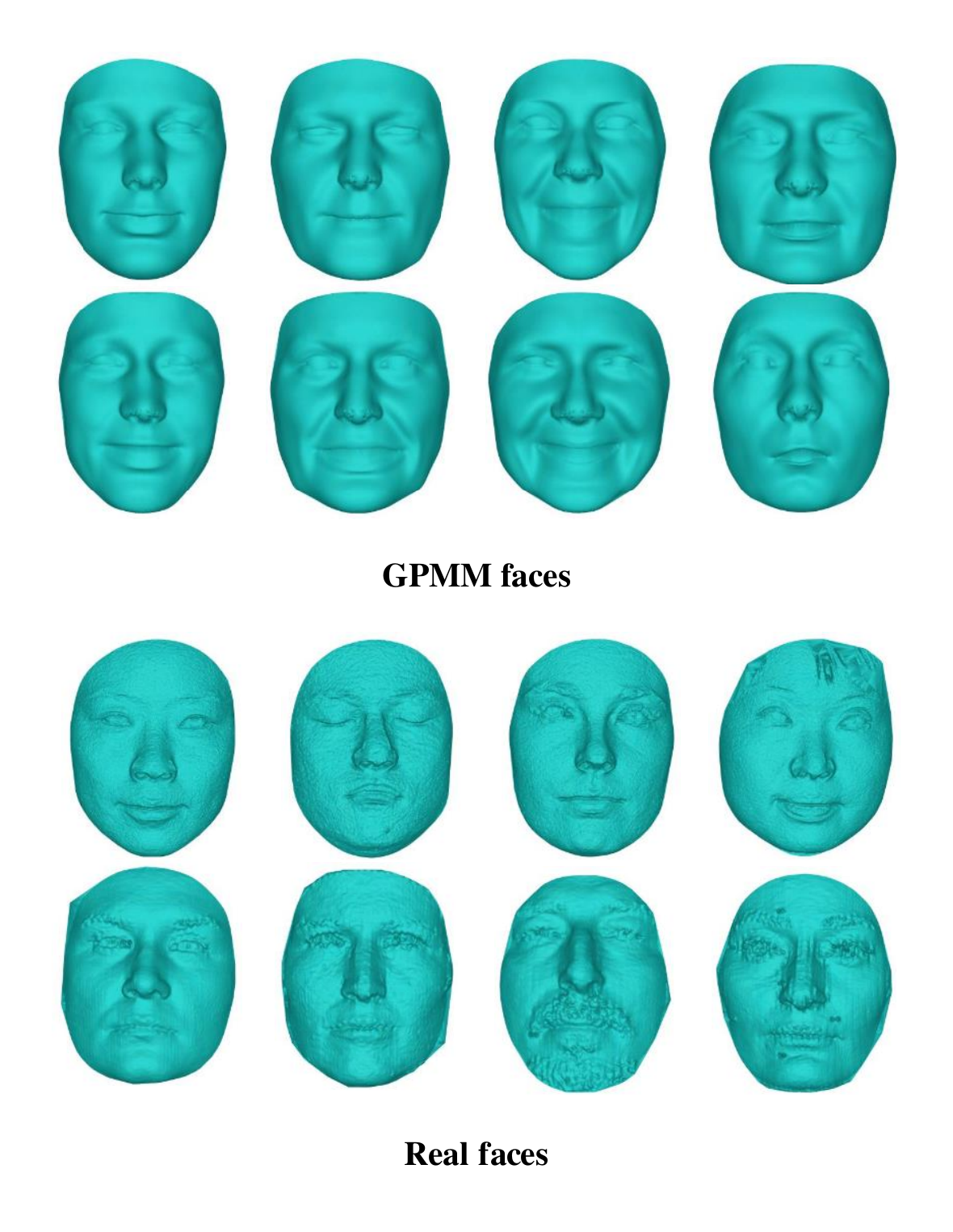}
	\end{center}
	\caption{GPMM faces vs Real faces. The first two columns are four GPMM identities with two different expressions. The last two columns are real faces from FRGCv2.0 and Bosphorus dataset.}
	\label{fig:faces}
\end{figure}

We totally generated 10k independent $\alpha_i$ for unique identities and 50 $\beta_j$ for expressions. Example facial scans synthesized from the generator are illustrated in the first two columns of Fig.~\ref{fig:faces}. Compared to the real faces show in the last two columns of Fig.~\ref{fig:faces}, the most distinct is the details in eyes, nose, mouths which will change according to the poses and expressions. According to these differences between the train and real datasets, we propose to apply different sample strategy to extract the same features. The sampling method based on curvatures will be introduced next.
\subsection{Curvature-aware sampling strategy}
In PointNet++, iterative farthest point sampling(FPS) strategy is used to choose the centroids subset based on the metric distance. Compared with random sampling, FPS has better coverage of the entire point set on given the same number of centroids.
However, 3D face point clouds in the strict sense do not belong to metric point set. Different from objects in ModelNet40 dataset, e.g. planes and chairs, face scans just have one surface and all points lie on it.

\begin{figure}[t]
	\begin{center}
		\includegraphics[width=1\linewidth]{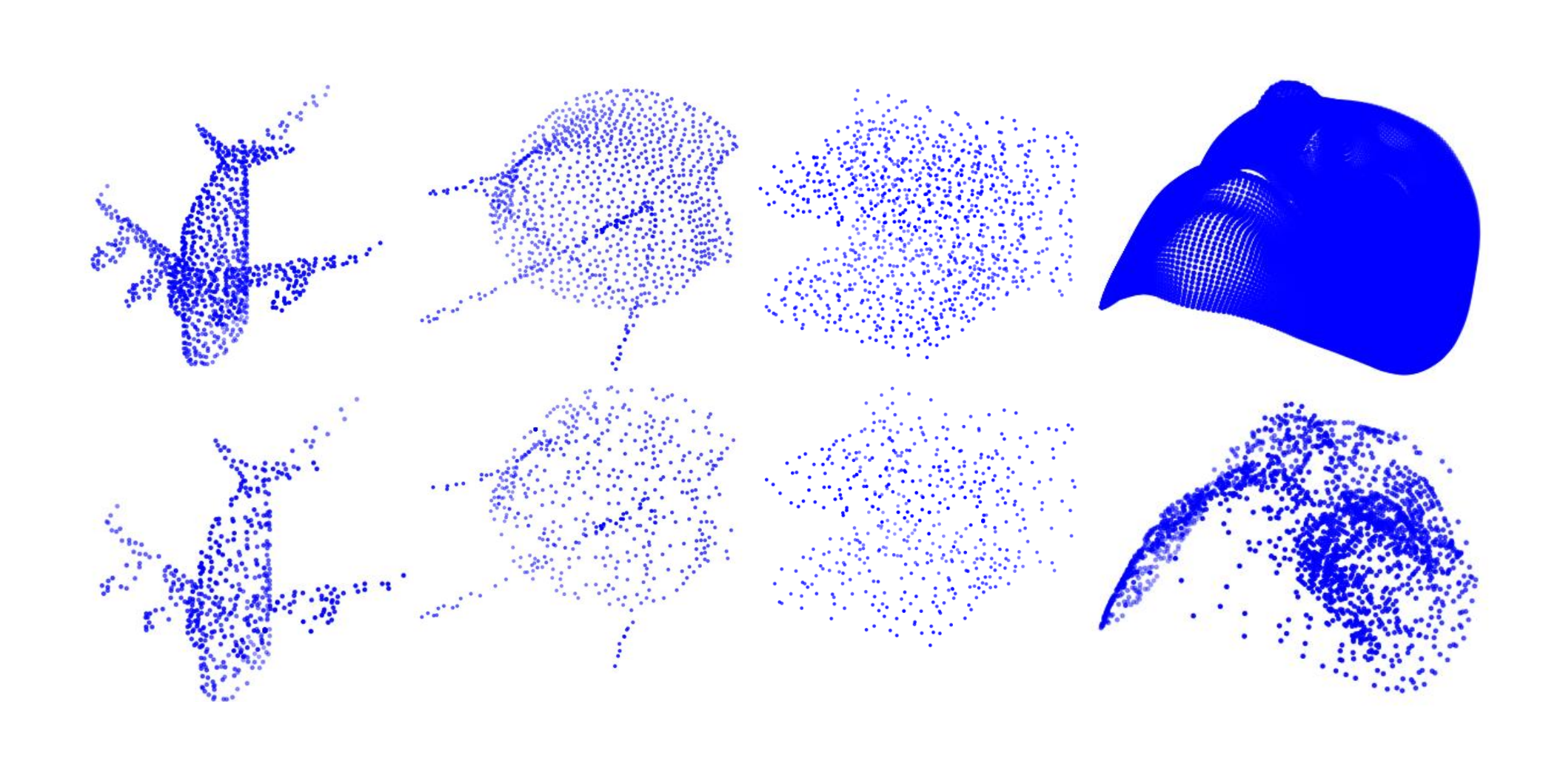}
	\end{center}
	\caption{FPS results on ModelNet objects and generated face. The top row are the original point clouds of plane, chair, desk and a face. The bottom row are the sampled results with 512 points.}
	\label{fig:fps}
\end{figure}

Using the FPS on planes, chairs and desks, some special points will be sampled out such as the corners of desk, the edges in plane wings and the corners in chair legs as see in Fig.~\ref{fig:fps}. But to face point clouds, the FPS result in the contour points which may only represent the face size and the noise or spark points which is not very useful for face representation. Only few points localize in the nose, eye area where prove to be more important in face recognition.

In FPS, the Euclidean distance is used to measure the relation between points. However, to faces, Euclidean distance is not a good choice because it is sensitive to miscellaneous and contour points, especially in the real face point clouds which is largely affected by the measure accuracy containing more noise points. So, sampling centroids on faces is different from the common objects. We need to modify the metric strategy to pay more attention to the feature points instead of the distant points. Inspired by the DGCNN\cite{wang2019dgcnn} and PointConv\cite{wu2019pointconv}, we combine the facial curvature with the metric distance to represent the facial feature distance. Using the modified distance, we can filter out the points with high curvature in a local region.

\begin{figure}[t]
	\begin{center}
		\includegraphics[width=1\linewidth]{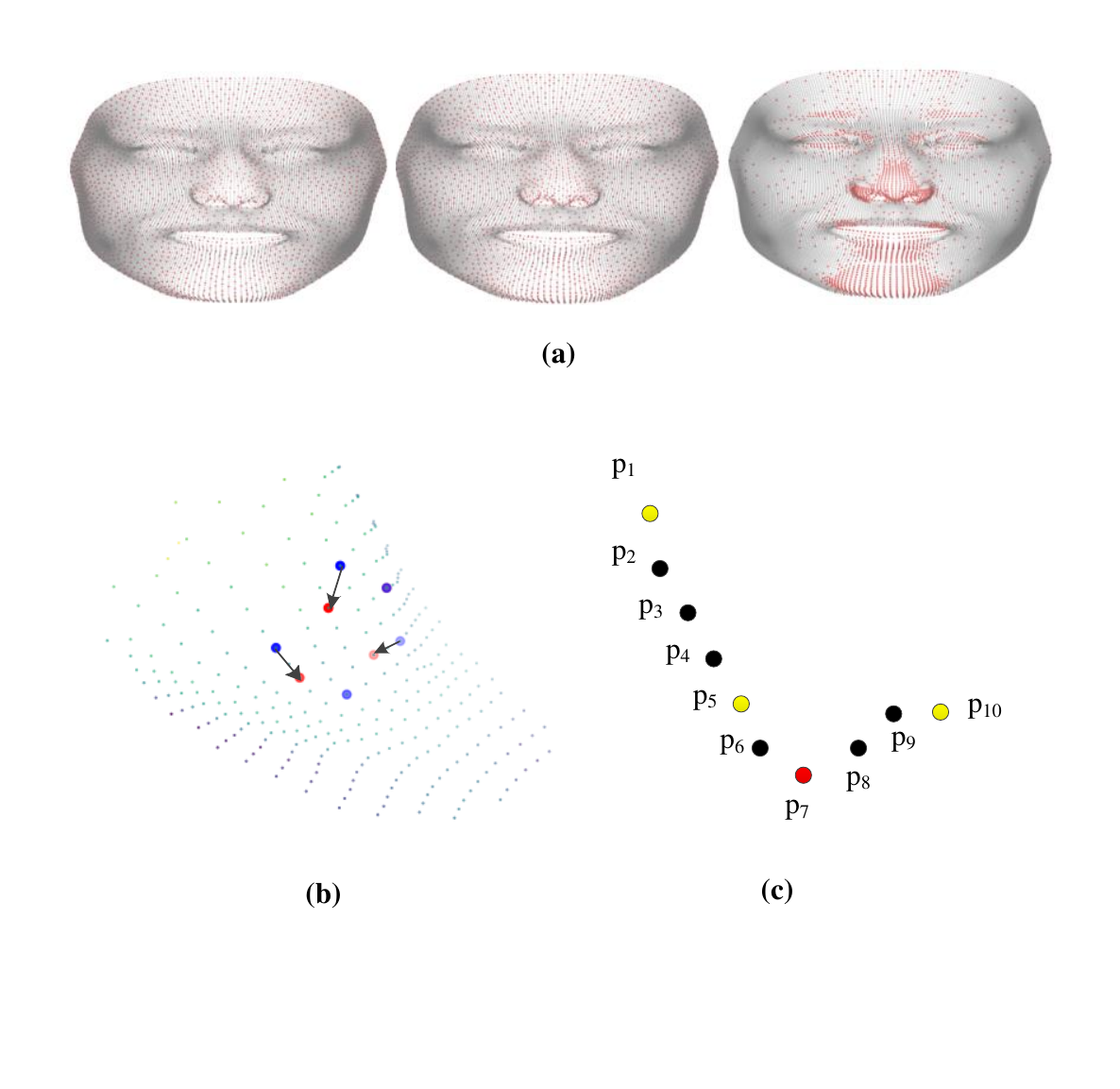}
	\end{center}
	\caption{Curvature-aware point sampling results on faces compared with FPS. (a) shows the CPS results in different $\lambda$ (results in red). From left to right is $\lambda=0$ (equal to FPS), $\lambda=0.1$ and $\lambda=1$. (b) A detail area about the curvature's contribution. The arrows from blue point to red is the effect of curvature. (c) A 2D illustration of the CPS.}
	\label{fig:ffs}
\end{figure}

Givn the input point clouds $P=\begin{Bmatrix}P_1,P_2,...,P_n\end{Bmatrix}$, the FPS result in the output subset $Q=\begin{Bmatrix}P_{q1},P_{q2},...,P_{qm}\end{Bmatrix}$. The aim is to search the point $P_{qi}$ in the remaining point set$\begin{Bmatrix}P-Q\end{Bmatrix}$to satisfy:
\begin{equation}
argmax \sum_{k=1}^{j-1} d(P_{qi},P_{qk})
\end{equation}
where, $d(P_{qi},P_{qk})$ computer the Euclidean distance between $P_{qi}$ and $P_{qk}$. Now, we fuse the Euclidean distance with facial curvature.
\begin{equation}d_c=d(P_{qi},P_{qk})\cdot C_{qi}^ \lambda\end{equation}
where the $C_{qi}$ represent the curvature of the point $P_{qi}$ and $\lambda$ is a hyper-parameter balance the curvature. In the log format
\begin{equation}log(d_c)=log(d(P_{qi},P_{qk})) +  \lambda C_{qi}\end{equation}
From the log view, we can see that curvature contributes to the distance by $\lambda$. Points with large curvature get large distance and is more likely to be chosen. But we can suppress it by a factor $\lambda$ other wise it will destroy the normal distribution of points. When $\lambda=0$, CPS equals to the original FPS. As show in Fig.~\ref{fig:ffs}(a), from left to right is the original FPS result on face, $\lambda=0.1$ CPS result and $\lambda=1$ CPS result. We can see that when $\lambda=1$ the sampling focus on some landmark points. In this case, the sampling result is extremely affected by the curvature which is not beneficial to extract features. We choose the best $\lambda$ according to experiments in Sec.~\ref{Sec:experiments}. In Fig.~\ref{fig:ffs}(b), we can see the details about the curvature function. The FPS's point(in red) will shit to the blue points according to curvature. A 2D illustration in Fig.~\ref{fig:ffs}(c) is more clearly. We use the FPS to choose 3 target points from 10 source points $\begin{Bmatrix}p_1,p_2,...,p_{10}\end{Bmatrix}$. The original FPS will choose $p_1$, $p_{10}$, $p_5$ in distance order. But we can clearly know that $p_7$ is more representative than $p_5$. By means of curvature, our sampling strategy choose the $p_7$ replace the $p_5$.

In addition, to avoid interference from outliers and edge points, we limit candidate regions for sampling points. Because the size of face is an inherent property, so we can use an absolute value and do not need to consider issues such as face symmetry and missing. Before CPS, we filter the points far from the noes tip according to the distance $r$. We should notice that the filered points are just excluded from the sample candidates and they are still in the neighbors contributing to the local features.
\subsection{Network Architecture}
Inspired by the success of PointNet++\cite{qi2017pointnet++:}, we propose to use the similar network structure to extract face representations. The proposed network is visualized in Fig.~\ref{fig:overview}.
Three set abstraction(SA) modules are used which contains sampling, grouping and MLP layers to extract local-global features. The first two SA focus on local feature with different receptive field and the last one aggregate global feature.
The proposed CPS is integrated in sampling layer to sample key points.
We change the ball radius in grouping layer to fit the face scale and density.

We take the global feature as the face embedding as proposed in \cite{Schroff_2015facenet}. The embeddings, represent discriminative face feature, can be used to calculate the cosine similarity between faces. If the distance with two scans larger than the given threshold, we consider the two scans belong to the same identity, vice versa.

We use angular loss\cite{liu2017sphereface} during the classification task training. The angular loss added a margin between different classes to enlarge the inter-class distance\cite{deng2018arcface,wen2016discriminative,wang2018cosface}.
\begin{equation}
\label{eq:angular}
L_{a}=-\frac{1}{N}\sum_{i=1}^{N}log\frac{e^{scos(m\theta _{y_i})}}{e^{scos(m\theta _{y_i})}+\sum_{j=1,y\neq y_i}^{C}e^{scos\theta _j}}
\end{equation}
where the bias is fixed to $0$. $\theta_{j}$ is the angle between the weight $W_j$ and the feature $x_i$. $s$ is the scale factor to control the convergence. $m$ is an additive angular margin penalty to simultaneously enhance the intra-class compactness and inter-class discrepancy.

After classification training, we fine tune our network on a few real faces using the triplet loss \cite{Schroff_2015facenet}. We change the metric distance in triplet loss function from Euclidean to cosine. Using triplet loss instead of softmax loss avoids optimizing a new fully-connected classify layer which is time inefficient. And to real databases, using triplet loss can converge quickly and fixable. 
\begin{equation}
\label{eq:softmax}
L_{t} =  \frac{1}{N}\sum_{i=1}^{N}(d(x_i^a,x_i^p)-d(x_i^a,x_i^n)+\alpha )
\end{equation}
where $d(\cdot )$ computer the cosine distance between two vectors. $x_i^a,x_i^p,x_i^n$ is the anchor, positive and negative samples respectively. $\alpha$ is the margin that is enforced between positive pairs and negative pairs.
\section{Experiments and results}
\label{Sec:experiments}
In this section, we first evaluate the affects of proposed sampling strategy and augment data for 3D faces. Massive experiments are conducted to choose the hyper-parameters. And then, we select the best hyper-parameter setting for face recognition on different 3D face database. 
\subsection{Implementation Details}
In both classification training and fine tuning, the input to our network is face scans with $28588$ points. Each point has 7 dimension features including 3 Euclidean coordinates $xyz$, corresponding normal vectors $n_xn_yn_z$ and curvature $c$. The normal vector and curvature are calculated by PCL \cite{Rusu_ICRA2011_PCL}.
We train the propose network in PyTorch \cite{paszke2017automatic}.
We use Adam \cite{kingma2014adam} optimizer and performs batch normalization for all layers excluding the last classification layer. The initial learning is set to 0.001 and reduced by a factor of 10 after every 20 epochs. And the weight decay is set to 0.5, which is decreased by 0.5 up to 0.99.
The network is trained for totally 60 epochs with batch size of 32 scans on a single NVIDIA GeForce GTX 1080TI GPU.
For fine tuning, learning is set to 1e-5 and we randomly sample triplet pairs from the part of Bosphorus and FRGCv2.0 database.
\subsection{Ablation Study}
There are three hyper-parameters: candidate region distance $r$, curvature factor $\lambda$ and train data numbers or classes $N$. We adopt the ablation study on each parameter. As the train data numbers affect the training time, the first two study in conducted on $N=500$ classes total 25,000 face scans to save the training time. And we last evaluate the train data classes vary from 500 classes to 10,000 classes.
We choose a subset from the Bosphorus database as the evaluate set $B_{eval}$.

\begin{figure}[t]
	\begin{center}
		\includegraphics[width=1\linewidth]{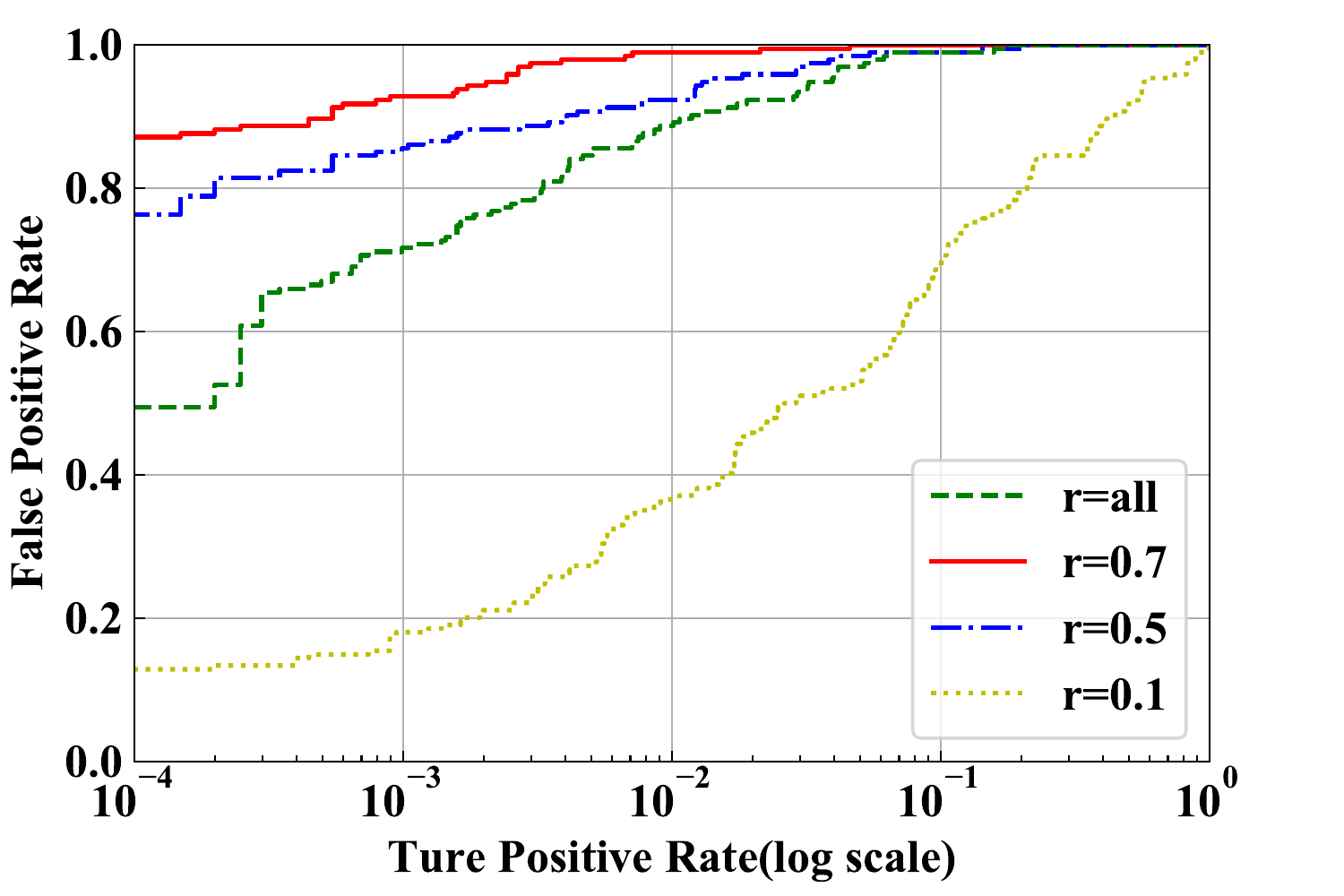}
	\end{center}
	\caption{ROC curves on $B_{eval}$ in different $r$.}
	\label{fig:p_result}
\end{figure}

\textbf{candidate region distance $r$}. We conduct open-set face recognition on the $B_{eval}$.  Fig.~\ref{fig:p_result} illustrates the importance of candidate region distance $r$. Either big $r$ or small $r$ is not suitable for 3D face data. $r=all$ equals the original FPS, we can see that, after limit the candidate in about $0.5\sim 0.7$ the ROC curves are higher and perform better. When $r$ cover all points, sampling result will contain more edge points which will decrease the recognition performance. When $r$ is small, sampling result only keeps a litter face area and loss some useful information. To balance the effect between redundant edge points and useful facial representation, we choose $r=0.7$ to get the best result and use for next experiments.

\begin{figure}[t]
	\begin{center}
		\includegraphics[width=1\linewidth]{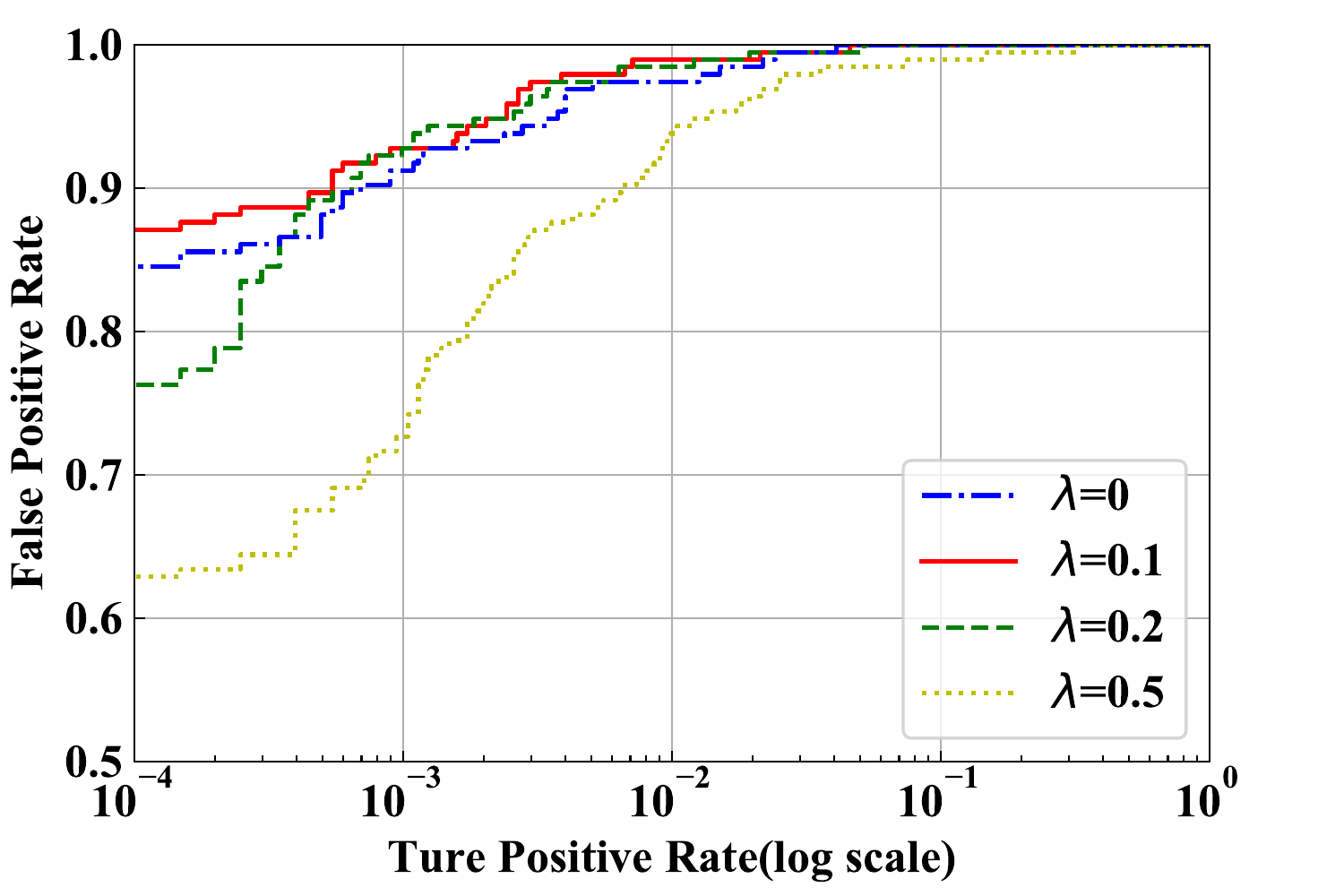}
	\end{center}
	\caption{ROC curves on $B_{eval}$ in different $\lambda$.}
	\label{fig:lambda_result}
\end{figure}

\textbf{curvature factor $\lambda$}. We also evaluate the effectiveness of curvature factor $\lambda$ as shown in Fig.~\ref{fig:lambda_result}. $\lambda=0$ equals to the original FPS and means curvature is not introduced into the sampling. when $\lambda$ is too big $\lambda=0.5$, as shown in Fig.~\ref{fig:ffs}, the sampling points will focus on the high-curvature points and loss the uniform, resulting in a decrease gap in recognition performance. We can see that a small curvature factor $\lambda=0.1$ is useful to slightly shift some points without destroying the uniform distribution.

\begin{figure}[t]
	\begin{center}
		\includegraphics[width=1\linewidth]{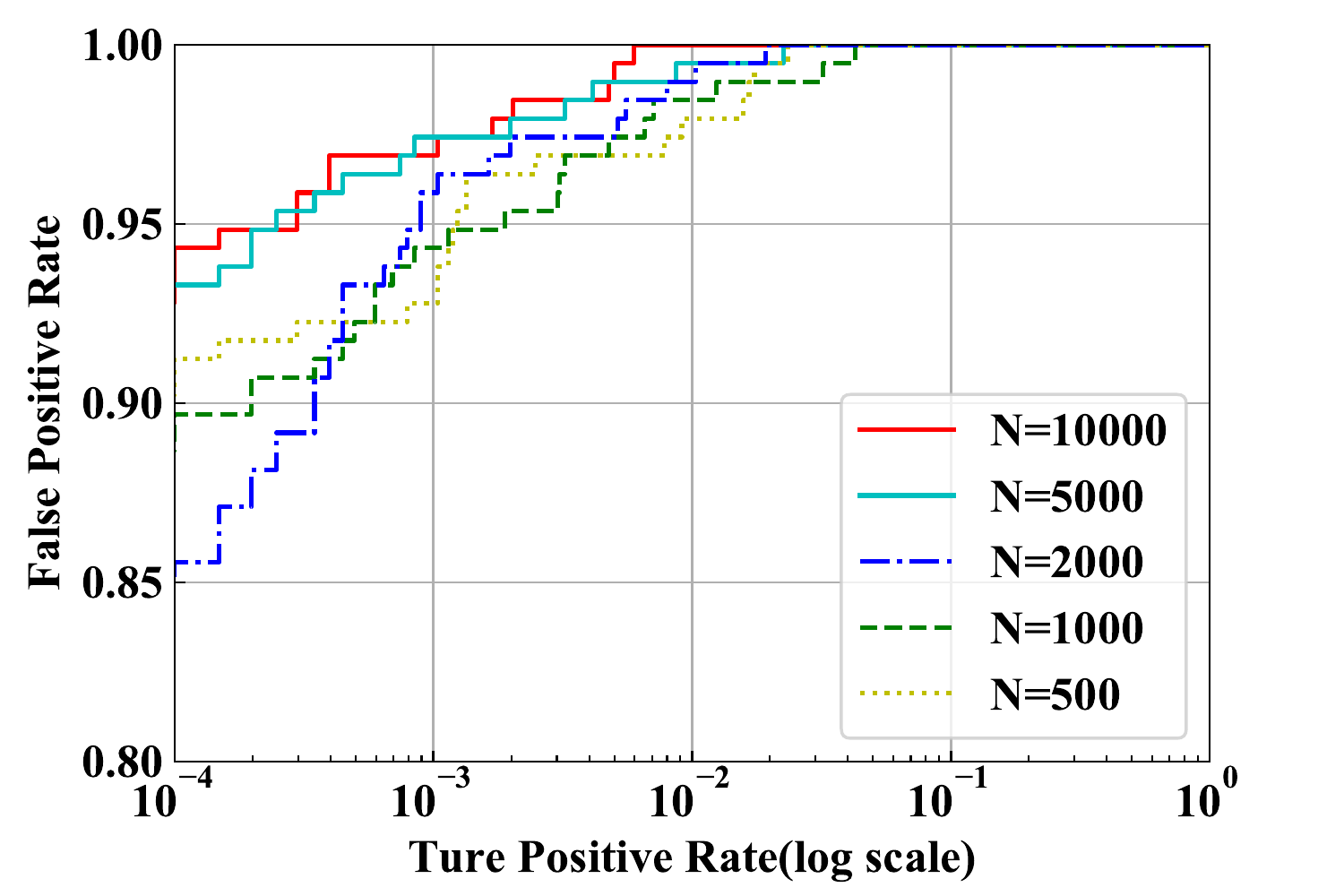}
	\end{center}
	\caption{ROC curves on $B_{eval}$ in different $N$.}
	\label{fig:N_result}
\end{figure}

\textbf{training classes $N$}. One of the advantage of our data augmentation method is that we can generate infinite number of training data. But too many data leads to the time inefficient. So we should achieve a balance between recognition performance and training time waste. As seen in Fig.~\ref{fig:N_result}, the more data network train, the higher recognition accuracy. However, the training time increases linearly with $N$. While $N=500$ needs 4 hours, $N=10,000$ waste 3 days. When $N$ is large, the training time required to increase the recognition accuracy by 0.1\% is several times higher than previous.
So we train our work on 10,000 classes in this paper. And if you need higher performance, you can train on bigger $N$.
\subsection{Generalization on FRGC and Borphous}
The proposed approach is also evaluated on FRGCv2.0 and Borphous databases for face identification under different expression settings.

\begin{figure}[t]
	\begin{center}
		\includegraphics[width=1\linewidth]{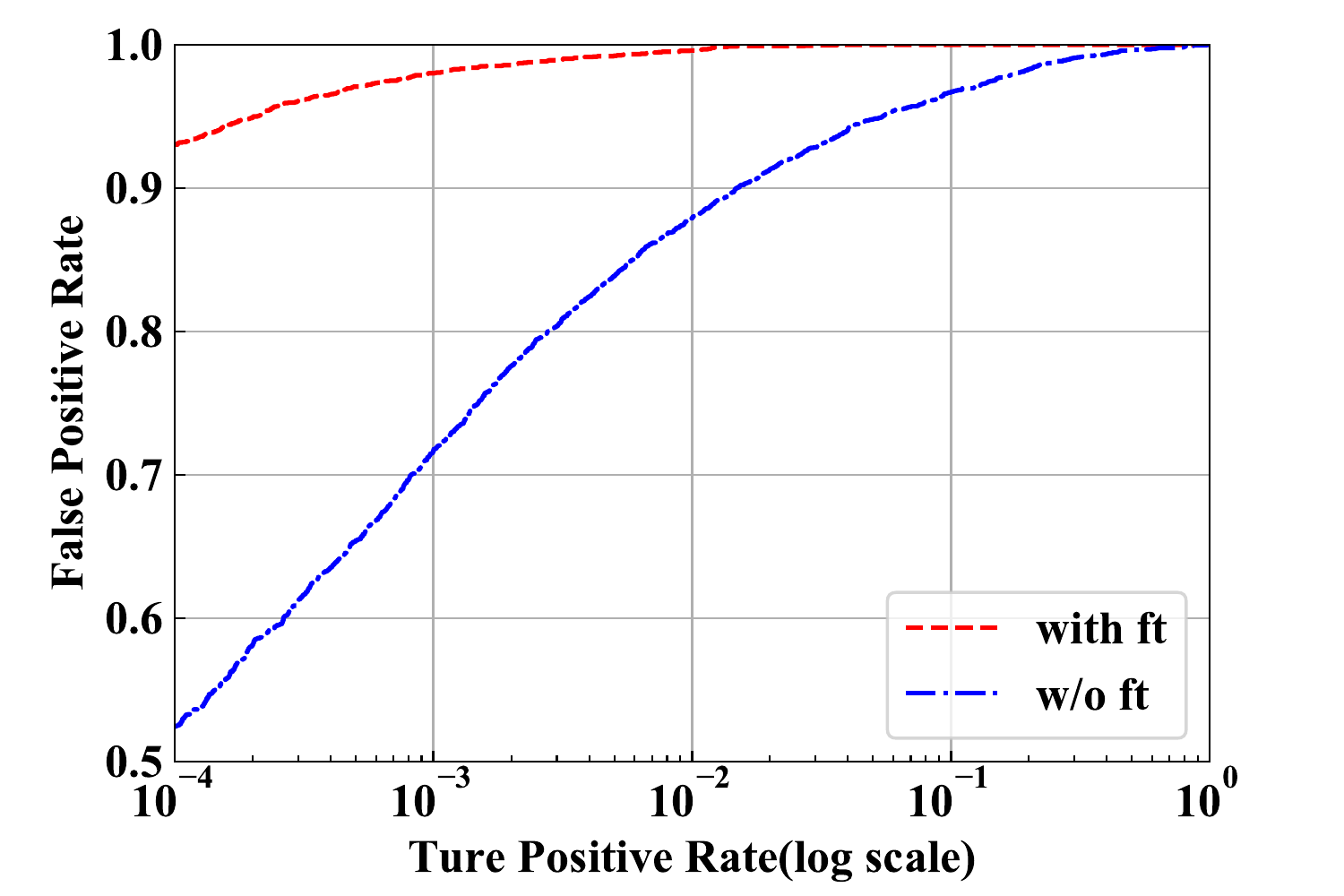}
	\end{center}
	\caption{ROC curves on FRGCv2.0 dataset with and without fine tuning.}
	\label{fig:frgc_result}
\end{figure}

For FRGCv2 database, the first scan of each identity is selected as gallery and the rest scans are treated as probes. We get 466 galleries and 3541 probes.
The rank-1 results are displayed in Tab.~\ref{tab:FRGC} compared with state-of-the-art methods.
We also train a PointNet++ network on the faces without our CPS. We can see that the raw PointNet++ perform far from the the-state-of-art methods, which demonstrate the effectiveness of our proposed CPS method.
Compared with FR3DNet\cite{Zulqarnain_Gilani_2018} which get a 97.06\% accuracy trained on over 3 million scans and 99.88\% after fine tuning, our network trained on 0.5 million faces still exists a gap. But we think the network trained without any real data has achieved our goal. Actually we can further enhance our network by using more training data or fine tuning on real faces. We have confirmed that more training data contributing to higher performance before. So we tempt to fine tune on galleries of FRGCv2.0 and our network shows a great promote which nearly equal to FR3DNet\cite{Zulqarnain_Gilani_2018}. We compare our ROCs with and without fine tuning in Fig.~\ref{fig:frgc_result}. There is a significant upgrade after fine tuning on real faces. This encourage us to first train on larger generated data to pre-train the network and then fine tune on a few real faces to get best recognition performance.

\begin{table}[]
	\caption{Comparison of the rank-1 on the FRGCv2.0 database.} 
	\label{tab:FRGC}
	\begin{center}
	\small
	\begin{tabular}{l|c}
				\hline
				Methods      								&  Rank-1 Accuracy	\\ \hline
				MMH\cite{mian2007efficient}					&  96.2				\\
				TPWCRC\cite{Lei2016A}						&  96.3			\\
				FGM\cite{Li2015Towards}						&  96.3			\\
				K3DM\cite{gilani2017dense}					&  98.50  			\\
				FR3DNet\cite{Zulqarnain_Gilani_2018}		&  97.06  			\\
				FR3DNet(ft)\cite{Zulqarnain_Gilani_2018}    &   \textbf{99.88} 	\\ \hline
				PointNet++   								&  72.55 			\\
				Our work 									&  92.74  			\\
				Our work(ft)     							&  \textbf{98.73} 	\\ \hline
	\end{tabular}
	\end{center}
\end{table}
For Bosphorus database, we follow the protocol in \cite{berretti2013matching} to evaluate the expression performance.
For each subject,the first neutral scan is included in the gallery set, whereas the remaining scans are divided into Neutral, expressive, Lower Face Action Unit (LFAU), Upper Face Action Unit (UFAU) and Combined Action Unit (CAU).
The results are displayed in Tab.~\ref{tab:Borphous}. We can see that our network trained only on generated data has overall performance close to the-state-of-art methods. This demonstrate that our recognition method generalize well for unseen identities. After fine tuning on the first 30 identities in Bosphorus database, our network outperforms 10\% than original one.
The FR3DNet in \cite{Zulqarnain_Gilani_2018} achieved 100\% after fine tuning which using large real faces.
\begin{table}[]
\caption{Comparison of the rank-1 rate(\%) on the Bosphorus database.} 
\label{tab:Borphous}
\begin{center}
	\small
	\begin{tabular}{l|c|c|c}
				\hline
				Probes      		&Berretti\cite{berretti2013matching}& Our work & Our work(ft)\\ \hline
				Neutral(194)		& 97.9 	 	& 100   & 100   \\
				Anger(71)			& 85.9 		& 81.69 & 94.34 \\
				Disgust(69)     	& 81.2	 	& 79.71 & 89.86 \\ 
				Fear(70)   			& 90.0		& 88.57 & 94.29 \\
				Happy(106) 			& 92.5	 	& 96.23 & 100 \\
				Sad(66) 			& 93.9		& 90.91 & 95.45 \\
				Surprise(71) 		& 91.5	 	& 95.77 & 100	 \\\hline
				LFAU(1549) 			& 96.5		& 92.51 & 97.03 \\
				UFAU(432) 			& 98.4	 	& 96.76 & 99.07 \\
				CAU(169) 			& 95.6		& 95.86 & 98.82 \\\hline
				Overall(2797)    	& 95.67   	& 93.38 & \textbf{97.50} \\\hline
\end{tabular}
\end{center}
\end{table}

\section{Conclusion}
In this paper, we try to present a new 3D face recognition approach based on point cloud.
To avoid the problem of lacking large training dataset, we propose a data-free method that GPMM is introduced to synthesize model faces to replace real faces in training procedure. After analyzing the distribution difference between synthesized and real faces, a face area constraint method to promote consistency of distribution.
A curvature-aware point sampling strategy is proposed to filter distinct face points and extract deep features.
The proposed method generalizes well to unseen faces and after fine tuning with real faces it outperforms the state-of-the-art face recognition algorithms in the 3D domain. Our network is also competitive in case of expression and pose challenges.
{\small
\bibliographystyle{ieee_fullname}
\bibliography{egbib}

\begin{thebibliography}{10}\itemsep=-1pt

\bibitem{abate20072d}
Andrea~F Abate, Michele Nappi, Daniele Riccio, and Gabriele Sabatino.
\newblock {2D and 3D} face recognition: A survey.
\newblock {\em Pattern Recognition Letters}, 28(14):1885--1906, 2007.

\bibitem{Soltani_2017_CVPR}
Amir Arsalan~Soltani, Haibin Huang, Jiajun Wu, Tejas~D. Kulkarni, and Joshua~B.
  Tenenbaum.
\newblock Synthesizing {3D} shapes via modeling multi-view depth maps and
  silhouettes with deep generative networks.
\newblock In {\em CVPR}, 2017.

\bibitem{berretti2013matching}
Stefano Berretti, Naoufel Werghi, Alberto Del~Bimbo, and Pietro Pala.
\newblock Matching {3D} face scans using interest points and local histogram
  descriptors.
\newblock {\em Computers \& Graphics}, 37(5):509--525, 2013.

\bibitem{blanz19993DMM}
Volker Blanz and Thomas Vetter.
\newblock A morphable model for the synthesis of {3D} faces.
\newblock In {\em SIGGRAPH}, pages 187--194, 1999.

\bibitem{bowyer2006a}
Kevin~W Bowyer, Kyong~I Chang, and Patrick~J Flynn.
\newblock A survey of approaches and challenges in {3D} and multi-modal {3D +
  2D} face recognition.
\newblock {\em Computer Vision and Image Understanding}, 101(1):1--15, 2006.

\bibitem{cao2018vggface2}
Qiong Cao, Li Shen, Weidi Xie, Omkar~M Parkhi, and Andrew Zisserman.
\newblock Vggface2: A dataset for recognising faces across pose and age.
\newblock In {\em FG}, pages 67--74, 2018.

\bibitem{charles2017pointnet:}
R~Qi Charles, Hao Su, Mo Kaichun, and Leonidas~J Guibas.
\newblock {PointNet}: Deep learning on point sets for {3D} classification and
  segmentation.
\newblock In {\em CVPR}, pages 77--85, 2017.

\bibitem{deng2018arcface}
Jiankang Deng, Jia Guo, Niannan Xue, and Stefanos Zafeiriou.
\newblock {ArcFace}: Additive angular margin loss for deep face recognition,
  2018.

\bibitem{Dou_2017_CVPR}
Pengfei Dou, Shishir~K. Shah, and Ioannis~A. Kakadiaris.
\newblock End-to-end {3D} face reconstruction with deep neural networks.
\newblock In {\em CVPR}, 2017.

\bibitem{drira20133d}
Hassen Drira, Boulbaba~Ben Amor, Anuj Srivastava, Mohamed Daoudi, and Rim
  Slama.
\newblock {3D} face recognition under expressions, occlusions, and pose
  variations.
\newblock {\em IEEE Transactions on Pattern Analysis and Machine Intelligence},
  35(9):2270--2283, 2013.

\bibitem{faltemier2007ND-2006}
Timothy~C Faltemier, Kevin~W Bowyer, and Patrick~J Flynn.
\newblock Using a multi-instance enrollment representation to improve {3D} face
  recognition.
\newblock In {\em ICB}, pages 1--6, 2007.

\bibitem{Feng_2018_CVPR}
Yifan Feng, Zizhao Zhang, Xibin Zhao, Rongrong Ji, and Yue Gao.
\newblock Gvcnn: Group-view convolutional neural networks for 3d shape
  recognition.
\newblock In {\em CVPR}, 2018.

\bibitem{Gecer_2019_GANFIT}
Baris Gecer, Stylianos Ploumpis, Irene Kotsia, and Stefanos Zafeiriou.
\newblock {GANFIT}: Generative adversarial network fitting for high fidelity 3d
  face reconstruction.
\newblock In {\em CVPR}, 2019.

\bibitem{gilani2017dense}
Syed~Zulqarnain Gilani, Ajmal Mian, Faisal Shafait, and Ian Reid.
\newblock Dense {3D} face correspondence.
\newblock {\em IEEE Transactions on Pattern Analysis and Machine Intelligence},
  40(7):1584--1598, 2017.

\bibitem{gupta2010anthropometric}
Shalini Gupta, Mia~K Markey, and Alan~C Bovik.
\newblock Anthropometric {3D} face recognition.
\newblock {\em International Journal of Computer Vision}, 90(3):331--349, 2010.

\bibitem{Huibin2014Expression}
L.~I. Huibin, D.~I. Huang, MORVAN, JeanMarie, Liming Chen, and Yunhong Wang.
\newblock Expression-robust {3D} face recognition via weighted sparse
  representation of multi-scale and multi-component local normal patterns.
\newblock {\em Neurocomputing}, 133(14):179--193, 2014.

\bibitem{jackson2017large}
Aaron~S Jackson, Adrian Bulat, Vasileios Argyriou, and Georgios Tzimiropoulos.
\newblock Large pose {3D} face reconstruction from a single image via direct
  volumetric {CNN} regression.
\newblock In {\em ICCV}, pages 1031--1039, 2017.

\bibitem{Jian2016Research}
Zhang Jian, Zhenjie Hou, Zhuoran Wu, Yongkang Chen, and Weikang Li.
\newblock Research of {3D} face recognition algorithm based on deep learning
  stacked denoising autoencoder theory.
\newblock In {\em ICCSN}, pages 663--667, 2016.

\bibitem{kim2017deep}
Donghyun Kim, Matthias Hernandez, Jongmoo Choi, and Gerard~G Medioni.
\newblock Deep {3D} face identification.
\newblock {\em International Journal of Central Banking}, pages 133--142, 2017.

\bibitem{kingma2014adam}
Diederik~P Kingma and Jimmy Ba.
\newblock Adam: A method for stochastic optimization.
\newblock In {\em ICLR}, 2015.

\bibitem{Learned2016LFW}
Erik Learned-Miller, Gary~B. Huang, Aruni Roychowdhury, Haoxiang Li, and Hua
  Gang.
\newblock Labeled faces in the wild: A survey.
\newblock In {\em Advances in Face Detection and Facial Image Analysis}, pages
  189--248, 2016.

\bibitem{Lei2016A}
Yinjie Lei, Yulan Guo, Munawar Hayat, Mohammed Bennamoun, and Xinzhi Zhou.
\newblock A two-phase weighted collaborative representation for {3D} partial
  face recognition with single sample.
\newblock {\em Pattern Recognition}, 52(C):218--237, 2016.

\bibitem{Li2015Towards}
Huibin Li, Di Huang, Jean~Marie Morvan, Yunhong Wang, and Liming Chen.
\newblock Towards {3D} face recognition in the real: A registration-free
  approach using fine-grained matching of {3D} keypoint descriptors.
\newblock {\em International Journal of Computer Vision}, 113(2):128--142,
  2015.

\bibitem{li2018pointcnn}
Yangyan Li, Rui Bu, Mingchao Sun, Wei Wu, Xinhan Di, and Baoquan Chen.
\newblock {PointCNN}: Convolution on $\mathcal{X}$-transformed points.
\newblock In {\em NeurIPS}, 2018.

\bibitem{liu2017sphereface}
Weiyang Liu, Yandong Wen, Zhiding Yu, Ming Li, Bhiksha Raj, and Le Song.
\newblock Sphereface: Deep hypersphere embedding for face recognition.
\newblock In {\em CVPR}, pages 212--220, 2017.

\bibitem{luthi2017GPMM}
Marcel L{\"u}thi, Thomas Gerig, Christoph Jud, and Thomas Vetter.
\newblock Gaussian process morphable models.
\newblock {\em IEEE Transactions on Pattern Analysis and Machine Intelligence},
  40(8):1860--1873, 2017.

\bibitem{mian2007efficient}
Ajmal Mian, Mohammed Bennamoun, and Robyn Owens.
\newblock An efficient multimodal {2D-3D} hybrid approach to automatic face
  recognition.
\newblock {\em IEEE Transactions on Pattern Analysis and Machine Intelligence},
  29(11):1927--1943, 2007.

\bibitem{parkhi2015VGGface}
Omkar~M Parkhi, Andrea Vedaldi, and Andrew Zisserman.
\newblock Deep face recognition.
\newblock In {\em BMVC}, pages 6--18, 2015.

\bibitem{paszke2017automatic}
Adam Paszke, Sam Gross, Soumith Chintala, Gregory Chanan, Edward Yang, Zachary
  DeVito, Zeming Lin, Alban Desmaison, Luca Antiga, and Adam Lerer.
\newblock Automatic differentiation in {PyTorch}.
\newblock In {\em NIPS Autodiff Workshop}, 2017.

\bibitem{Patil20153}
Hemprasad Patil, Ashwin Kothari, and Kishor Bhurchandi.
\newblock {3-D} face recognition: features, databases, algorithms and
  challenges.
\newblock {\em Artificial Intelligence Review}, 44(3):1--49, 2015.

\bibitem{phillips2005FRGCv2}
P~Jonathon Phillips, Patrick~J Flynn, Todd Scruggs, Kevin~W Bowyer, Jin Chang,
  Kevin Hoffman, Joe Marques, Jaesik Min, and William Worek.
\newblock Overview of the face recognition grand challenge.
\newblock In {\em CVPR}, pages 947--954, 2005.

\bibitem{Qi_2016_CVPR}
Charles~R. Qi, Hao Su, Matthias Niessner, Angela Dai, Mengyuan Yan, and
  Leonidas~J. Guibas.
\newblock Volumetric and multi-view cnns for object classification on {3D}
  data.
\newblock In {\em CVPR}, 2016.

\bibitem{qi2017pointnet++:}
Charles~Ruizhongtai Qi, Li Yi, Hao Su, and Leonidas~J Guibas.
\newblock {PointNet++}: Deep hierarchical feature learning on point sets in a
  metric space.
\newblock In {\em NeurPIS}, pages 5099--5108, 2017.

\bibitem{Rusu_ICRA2011_PCL}
Radu~Bogdan Rusu and Steve Cousins.
\newblock {3D is here: Point Cloud Library (PCL)}.
\newblock In {\em ICRA}, 2011.

\bibitem{savran2008bosphorus}
Arman Savran, Ne{\c{s}}e Aly{\"u}z, Hamdi Dibeklio{\u{g}}lu, Oya
  {\c{C}}eliktutan, Berk G{\"o}kberk, B{\"u}lent Sankur, and Lale Akarun.
\newblock Bosphorus database for {3D} face analysis.
\newblock In {\em European Workshop on Biometrics and Identity Management},
  pages 47--56. Springer, 2008.

\bibitem{Schroff_2015facenet}
Florian Schroff, Dmitry Kalenichenko, and James Philbin.
\newblock {FaceNet}: A unified embedding for face recognition and clustering.
\newblock In {\em CVPR}, 2015.

\bibitem{Spreeuwers2011Fast}
Luuk Spreeuwers.
\newblock Fast and accurate {3D} face recognition.
\newblock {\em International Journal of Computer Vision}, 93(3):389--414, 2011.

\bibitem{wang2018SpecGNN}
Chu Wang, Babak Samari, and Kaleem Siddiqi.
\newblock Local spectral graph convolution for point set feature learning.
\newblock In {\em ECCV}, pages 52--66, 2018.

\bibitem{wang2018cosface}
Hao Wang, Yitong Wang, Zheng Zhou, Xing Ji, Dihong Gong, Jingchao Zhou, Zhifeng
  Li, and Wei Liu.
\newblock Cosface: Large margin cosine loss for deep face recognition.
\newblock In {\em CVPR}, pages 5265--5274, 2018.

\bibitem{wang2019dgcnn}
Yue Wang, Yongbin Sun, Ziwei Liu, Sanjay~E Sarma, Michael~M Bronstein, and
  Justin~M Solomon.
\newblock Dynamic graph {CNN} for learning on point clouds.
\newblock {\em ACM Transactions on Graphics}, 2019.

\bibitem{wei20193d}
Huawei Wei, Shuang Liang, and Yichen Wei.
\newblock {3D} dense face alignment via graph convolution networks, 2019.

\bibitem{wen2016discriminative}
Yandong Wen, Kaipeng Zhang, Zhifeng Li, and Yu Qiao.
\newblock A discriminative feature learning approach for deep face recognition.
\newblock In {\em ECCV}, pages 499--515, 2016.

\bibitem{wu2019pointconv}
Wenxuan Wu, Zhongang Qi, and Li Fuxin.
\newblock {PointConv}: Deep convolutional networks on {3D} point clouds.
\newblock In {\em CVPR}, pages 9621--9630, 2019.

\bibitem{wu2015modelnet}
Zhirong Wu, Shuran Song, Aditya Khosla, Fisher Yu, Linguang Zhang, Xiaoou Tang,
  and Jianxiong Xiao.
\newblock {3D} shapenets: A deep representation for volumetric shapes.
\newblock In {\em CVPR}, pages 1912--1920, 2015.

\bibitem{xu2006casia-3d}
Chenghua Xu, Tieniu Tan, Stan Li, Yunhong Wang, and Cheng Zhong.
\newblock Learning effective intrinsic features to boost {3D}-based face
  recognition.
\newblock In {\em ECCV}, pages 416--427, 2006.

\bibitem{xu2018spidercnn}
Yifan Xu, Tianqi Fan, Mingye Xu, Long Zeng, and Yu Qiao.
\newblock {SpiderCNN}: Deep learning on net sets with parameterized
  convolutional filters.
\newblock In {\em ECCV}, pages 87--102, 2018.

\bibitem{yi2014CASIA}
Dong Yi, Zhen Lei, Shengcai Liao, and Stan~Z. Li.
\newblock Learning face representation from scratch, 2014.

\bibitem{Yi_2019_MMFace}
Hongwei Yi, Chen Li, Qiong Cao, Xiaoyong Shen, Sheng Li, Guoping Wang, and
  Yu-Wing Tai.
\newblock {MMFace}: A multi-metric regression network for unconstrained face
  reconstruction.
\newblock In {\em CVPR}, 2019.

\bibitem{yin2006BU3DFE}
Lijun Yin, Xiaozhou Wei, Yi Sun, Jun Wang, and Matthew~J Rosato.
\newblock A {3D} facial expression database for facial behavior research.
\newblock In {\em FG}, pages 211--216, 2006.

\bibitem{Yuyi2019Sparse}
Yi {Yu}, Feipeng {Da}, and Yifan {Guo}.
\newblock Sparse icp with resampling and denoising for {3D} face verification.
\newblock {\em IEEE Transactions on Information Forensics and Security},
  14(7):1917--1927, 2019.

\bibitem{Zulqarnain_Gilani_2018}
Syed Zulqarnain~Gilani and Ajmal Mian.
\newblock Learning from millions of {3D} scans for large-scale {3D} face
  recognition.
\newblock In {\em CVPR}, 2018.

\end{thebibliography}
}

\end{document}